%% file: main.tex
\documentclass[runningheads]{llncs}

\usepackage[final,year=2026,ID=*****]{eccv}

\usepackage{eccvabbrv}
\usepackage{amsmath,amssymb,amsfonts}
\usepackage{comment}
\usepackage{makecell}
\usepackage{booktabs}
\usepackage{multirow}
\usepackage{tabularx}
\usepackage{array}
\usepackage{graphicx}
\usepackage{ulem}
\usepackage[table]{xcolor}
\usepackage{pgfplots}
\pgfplotsset{compat=1.18}
\usepgfplotslibrary{groupplots}
\usepackage{subcaption}

\usepackage{amssymb}% http://ctan.org/pkg/amssymb
\usepackage{pifont}% http://ctan.org/pkg/pifont
\newcommand{\xmark}{\ding{55}}%

\newcommand{\blfootnote}[1]{%
  \begingroup
  \renewcommand{\thefootnote}{}%
  \NoHyper\footnote{#1}\endNoHyper%
  \addtocounter{footnote}{-1}%
  \endgroup
}

\usepackage{hyperref}
\usepackage{orcidlink}

\newcommand{\lctim}{\textsc{LC-TIM}}

\newcommand{\best}[1]{\textbf{#1}}
\newcommand{\second}[1]{\uline{#1}}
\newcolumntype{Z}{>{\centering\arraybackslash}X}
\definecolor{tablelightblue}{rgb}{0.90, 0.94, 0.985} % More transparent blue (simulated as light blue

\begin{document}

\title{Locally Consistent Transductive Information
Maximization for Few-Shot Remote Sensing Scene Classification}

\titlerunning{LC-TIM for RS Few-Shot Scene Classification}

% TODO FINAL: replace with actual authors for camera-ready
\author{Karim El Khoury$^{1,*}$, Benoît Gérin$^{1,*}$, Benoît Macq$^{1}$, and Christophe De Vleeschouwer$^{1}$}
\authorrunning{El Khoury et al.}
\institute{$^{1}$ICTEAM, UCLouvain, Louvain-la-Neuve, Belgium\blfootnote{*Denotes equal contribution.}
\\ \texttt{karim.elkhoury@uclouvain.be - benoit.gerin@uclouvain.be}}

\maketitle

% ---------------------------------------------------------------
\begin{abstract}

Remote sensing scene classification is increasingly relying on foundation models pre-trained on large-scale Earth-observation data. Moreover, transductive inference, which exploits the collective statistical structure of the entire unlabeled query set, appears to naturally match remote sensing pipelines where large images are routinely split into patches and inferred as a batch. In this work, we introduce LC-TIM (Locally Consistent Transductive Information Maximization), which extends the state-of-the-art Transductive Information Maximization for Few-Shot CLIP (TIM++) objective with a local consistency regularizer that enforces prediction agreement between each query sample and its $\kappa$ nearest feature-space neighbors. The regularizer enters as a single multiplicative factor in the closed-form $q$-update, adding negligible computational overhead. We further propose a multi-source extension that fuses the affinity graph from multiple remote sensing foundation model, further boosting classification accuracy. To assess these methods, we establish the first comprehensive, open-source benchmark for transductive few-shot RS scene classification, evaluating LP++, TransCLIP, TIM++, and LC-TIM across ten diverse datasets, two remote sensing vision-language models, and across various few-shot settings. Our experiments show that transductive methods consistently outperform zero-shot baselines, and that LC-TIM achieves state-of-the-art accuracy, with the largest gains in the low-shot regime where neighborhood cues are most informative. Code is publicly available at: \url{https://github.com/elkhouryk/LC-TIM}.

\keywords{Transductive inference \and Few-shot learning \and Remote sensing \and Vision-language models \and Scene classification}
\end{abstract}

\input{sections/1-intro}

\input{sections/2-RW}

\input{sections/3-Method}

\input{sections/4-Experiments}

\input{sections/5-Conclusion}

\section*{Acknowledgements}
Benoît Gérin is funded by ARIAC (Walloon region grant No. 2010235) and Christophe De
Vleeschouwer is funded by the Fonds National de la Recherche Scientifique (FNRS).

\newpage

% ---------------------------------------------------------------
\bibliographystyle{splncs04}
\bibliography{main}

\end{document}

%% file: sections/1-intro.tex
% ---------------------------------------------------------------
\section{Introduction}
\label{sec:intro}

%Motivate RS for few-shot classfication
Remote sensing (RS) imagery drives vital time-critical applications such as environmental monitoring, precision agriculture, and emergency disaster response~\cite{yuan2020deep,phang2023satellite,streamlined2024fast}. Underpinning these applications is scene classification, the process of rapidly assigning semantic labels to satellite imagery. While remote sensing platforms continuously collect petabytes of data, utilizing this information is bottlenecked by traditional supervised classifiers that rely on massive labeled datasets. Because manual annotation is expensive and slow, standard models cannot be deployed quickly in urgent, evolving scenarios. This operational bottleneck creates an urgent need for few-shot scene classification, enabling models to accurately classify categories using only a handful of labeled examples~\cite{elkhoury2026few}.

%RS fondation models
To address these data bottlenecks, RS foundation models have emerged as a powerful alternative. Vision Language Models (VLMs) like CLIP~\cite{radford2021learning} leverage large scale contrastive pre-training on image text pairs to enable zero-shot classification using simplified generic text prompts such as \texttt{"a centered satellite photo of \{class\}"}. To bridge the domain gap, the RS community has developed specific VLMs such as GeoRSCLIP~\cite{zhang2024rs5m}, RemoteCLIP~\cite{liu2024remoteclip}, and SkyCLIP~\cite{wang2024skyscript} which are trained on curated image-caption datasets, yielding significantly better zero-shot accuracy on standard RS scene classification benchmarks~\cite{el2025enhancing}. Alongside language supervised models, self supervised vision encoders pre-trained directly on satellite imagery, most notably the RS specific DINOv3~\cite{simeoni2025dinov3} trained on millions of RS patches, provide rich geometric and semantic representations. However, even the strongest RS foundation models leave noticeable performance gaps when encountering fine grained or spectrally atypical scene classes. This challenge directly motivates few-shot classification, using a handful of labeled support examples to adapt these powerful representations to complex target domains, with dedicated few-shot classification benchmarks for RSVLMs being released in the past year~\cite{elkhoury2026few}.

%transductive few-shot motivation
Despite the promise of these recent benchmarks, they predominantly evaluate few-shot performance under an \textit{inductive inference} paradigm, where predictions are made for each unlabeled query image in isolation. This strict inductive assumption overlooks a powerful alternative approach: \textit{transductive few-shot adaptation}. Unlike inductive methods, transductive learning optimizes decision boundaries by leveraging the joint distribution and collective statistical structure of the entire unlabeled query set simultaneously. This paradigm is particularly compelling for remote sensing applications, where large-scale images are routinely divided into separate patches and inferred concurrently as a collective batch. Exploiting the intrinsic data structure and shared distribution of these co-occurring patches offers an intuitive mechanism to improve classification consistency. While these transductive inference techniques have already proven to yield strong zero-shot classification performance within remote sensing applications~\cite{el2025enhancing}, no prior work has yet explored their few-shot adaptation potential to bridge the domain gaps inherent to RS foundation models. Among recent works, \emph{Transductive Information Maximization for Few-Shot CLIP}, known as TIM++~\cite{li2026timplusplus}, stands out as one of the top-performing frameworks. It works by maximizing the mutual information between query features and predicted labels while regularizing the posterior distribution via zero-shot cross-modal priors. However, TIM++ optimizes predictions globally and lacks explicit local consistency constraints within the feature space, meaning it overlooks the fine-grained data structure among nearby query samples.

%Contributions
\noindent\textbf{Contributions.} In this work, we bridge this gap by introducing a novel local consistency term directly into the TIM++ objective function, thereby outperforming existing transductive approaches. Specifically:
\begin{itemize}
\item We propose \lctim{} (\emph{Locally Consistent Transductive Information Maximization}), which extends TIM++ via a local consistency regularizer that enforces prediction agreement between each query sample and its $\kappa$ nearest feature-space neighbors. This regularizer enters as a single extra multiplicative factor in the closed-form $\mathbf{q}$-update of TIM++, requiring a negligible computational overhead.
\item We extend \lctim{} to incorporate additional RS foundation models. Concretely, we augment the RSVLM-based neighborhood affinity term with DINOv3 satellite patch embeddings, providing an orthogonal structural prior that yields additional performance gains.
\item We establish the first comprehensive and open-source benchmark for transductive few-shot learning in remote sensing by evaluating prominent frameworks, including LP++~\cite{huang2024lpplusplus}, TransCLIP~\cite{zanella2024boosting}, TIM++~\cite{li2026timplusplus}, and \lctim{} across ten diverse datasets. This study demonstrates that \lctim{} consistently outperforms alternative transductive methods even when encountering severe remote sensing domain shifts.
\end{itemize}

%% file: sections/2-RW.tex
\section{Related Work}
\label{sec:related}

\subsection{RS Foundation Models}
\label{sec:related_fm}

The development of large RS image-text datasets has catalyzed rapid progress in RS foundation models. RemoteCLIP fine-tunes CLIP on a curated RS captioning and visual question-answering corpus, demonstrating strong generalization across RS scene classification and retrieval benchmarks~\cite{liu2024remoteclip}. SkyCLIP~\cite{wang2024skyscript} builds SkyScript, a semantically diverse dataset of 2.6\,M RS image-text pairs linked to geographic and categorical metadata, enabling fine-grained, attribute-aware zero-shot classification. GeoRSCLIP~\cite{zhang2024rs5m} scales this further with RS5M, a 5-million-pair RS image caption dataset assembled by filtering publicly available satellite imagery, and achieves state-of-the-art zero-shot performance on most RS classification benchmarks. A structured evaluation of these RSVLMs across ten RS datasets under zero-shot and inductive few-shot conditions was recently provided in~\cite{elkhoury2026few}, revealing that zero-shot accuracy is not always a reliable predictor of inductive few-shot adaptation performance and motivating more careful benchmarking. Beyond language-supervised VLMs, purely visual self-supervised RS encoders offer a highly complementary approach. For instance, the satellite variant of DINOv3~\cite{simeoni2025dinov3} is trained on $493\text{~M}$ satellite image patches via a self-distillation objective, generating patch embeddings that capture intricate texture, structural, and spectral details without requiring text supervision. Because these representations reflect local appearance statistics, they differ fundamentally from the global semantic features produced by CLIP-style encoders, rendering the two model families natural complements. Within our framework, these families play distinct, orthogonal roles: while the RSVLMs supply the frozen visual features and cross-modal zero-shot priors that anchor the transductive objective in \lctim{}, the DINOv3 extension incorporates patch embeddings to construct the affinity graph underlying our local consistency term. This design injects a robust structural prior into the optimization process.

\subsection{Transductive and Few-Shot Learning for VLMs}
\label{sec:related_fsl}

Transductive inference for classification dates back to Vapnik's statistical learning theory~\cite{vapnik1999overview}. When the unlabeled test set is available at inference time, its collective statistics provide free additional supervision. TIM (\emph{Transductive Information Maximization}) formulates transductive few-shot classification as the maximization of the mutual information between query features and predicted labels~\cite{boudiaf2020information}. It solves the resulting optimization via an alternating direction method with closed form updates. When applied to VLMs, transductive methods gain an additional degree of freedom because the frozen text encoder provides rich zero-shot class prototypes that can serve as regularizers. TIM++~\cite{li2026timplusplus} exploits this by augmenting the TIM objective with a KL divergence term that aligns query predictions to the CLIP zero-shot distribution. This prevents drift from pre-trained language knowledge when labeled data is scarce, achieving top tier few-shot performance on natural image benchmarks. Another recently proposed method, TransCLIP~\cite{zanella2024boosting}, models the query set as a Gaussian mixture and propagates initial pseudo labels through a Laplacian affinity graph built from CLIP feature cosine similarities. This approach is closely related in spirit to our neighborhood consistency term. Taking a different trajectory, LP++~\cite{huang2024lpplusplus} trains a linear probe on support features blended with a learnable CLIP zero-shot prior. Although it optimizes a parametric head, it leverages query distribution statistics at inference via the prior weighting, placing it alongside fully transductive approaches in practice. Despite these advances, none of these transductive methods has been evaluated on few-shot RS classification, with TransCLIP being the sole exception evaluated on zero-shot RS scene classification~\cite{el2025enhancing}. Our work fills this gap by establishing the first comprehensive transductive few-shot benchmark for RS scene classification and proposing the novel \lctim{} transductive approach.\\

%% file: sections/3-Method.tex
% ============================================================
% Method section
% \bm requires . If your template does not load
% it, the fallback below maps \bm to amsmath's \boldsymbol:
\providecommand{\bm}[1]{\boldsymbol{#1}}
% Define a placeholder macro for your method name so it can be
% renamed globally later:
\newcommand{\ours}{\textsc{Ours}\xspace} % <-- rename here
% ============================================================

\section{Method}
\label{sec:method}

We present our framework in five steps: the problem setup and notation (Sec.~\ref{sec:setup}), the TIM++ base objective (Sec.~\ref{sec:timpp}), our local consistency regularization (Sec.~\ref{sec:lc}), its multi-source extension fusing affinities from complementary encoders (Sec.~\ref{sec:multisource}), and the unified optimization procedure (Sec.~\ref{sec:optim}).

\subsection{Problem Setup and Notation}
\label{sec:setup}

Let $\theta_v(\cdot)$ and $\theta_t(\cdot)$ denote the frozen visual and text encoders of a pre-trained VLM such as CLIP~\cite{radford2021learning} or GeoRSCLIP~\cite{zhang2024rs5m}. For an input image $\bm{x}_i$ and a textual description $\bm{c}_k$ of class $k \in \{1,\dots,K\}$, the $\ell_2$-normalized embeddings are
\begin{equation}
    \bm{f}_i = \frac{\theta_v(\bm{x}_i)}{\lVert \theta_v(\bm{x}_i) \rVert_2} \in \mathbb{R}^d,
    \qquad
    \bm{t}_k = \frac{\theta_t(\bm{c}_k)}{\lVert \theta_t(\bm{c}_k) \rVert_2} \in \mathbb{R}^d .
    \label{eq:embeddings}
\end{equation}
The zero-shot predictor of the VLM assigns class probabilities via
\begin{equation}
    \hat{y}_{ik} = \frac{\exp\!\big(\eta\, \bm{f}_i^{\top} \bm{t}_k\big)}{\sum_{j=1}^{K} \exp\!\big(\eta\, \bm{f}_i^{\top} \bm{t}_j\big)},
    \label{eq:zeroshot}
\end{equation}
where $\eta$ is a fixed temperature scaling.\\

In the transductive few-shot setting, we are given a small labeled support set $\mathcal{S} = \{(\bm{x}_i, y_{ik})\}_{i \in \mathcal{S}}$, with $y_{ik} \in \{0,1\}$ indicating whether sample $i$ belongs to class $k$, and an unlabeled query set $\mathcal{Q} = \{\bm{x}_i\}_{i \in \mathcal{Q}}$ drawn from the same $K$ classes. Transductive methods process $\mathcal{Q}$ jointly, exploiting its collective statistics as an additional unsupervised signal. 

\subsection{Background: TIM++}
\label{sec:timpp}
TIM++~\cite{li2026timplusplus} transductively optimizes a soft classifier $\mathbf{W} = [\bm{w}_1, \dots, \bm{w}_K] \in \mathbb{R}^{d \times K}$, where each column $\bm{w}_k$ acts as the prototype of class $k$. The posterior probability of class $k$ for sample $i$ is
\begin{equation}
    p_{ik} = \frac{\exp\!\big(-\tfrac{\tau}{2} \lVert \bm{f}_i - \bm{w}_k \rVert^2\big)}{\sum_{j=1}^{K} \exp\!\big(-\tfrac{\tau}{2} \lVert \bm{f}_i - \bm{w}_j \rVert^2\big)},
    \label{eq:posterior}
\end{equation}
with $\tau > 0$ a temperature parameter.\\

Writing $\bm{p}_i = (p_{i1}, \dots, p_{iK})$ and $\hat{\bm{y}}_i = (\hat{y}_{i1}, \dots, \hat{y}_{iK})$, the TIM++ objective combines three complementary terms:
\begin{equation}
    \min_{\mathbf{W}} \;\;
    \underbrace{\lambda \, \mathrm{CE}(\mathbf{W}; \mathcal{S})}_{\text{support supervision}}
    \;-\; \underbrace{\hat{\mathcal{I}}_{\alpha}(X_{\mathcal{Q}}; Y_{\mathcal{Q}})}_{\text{mutual information}}
    \;+\; \underbrace{\gamma \, \mathcal{D}_{\mathrm{KL}}(\bm{p} \,\Vert\, \hat{\bm{y}})}_{\text{text alignment}} ,
    \label{eq:timpp}
\end{equation}
where $\lambda > 0$, $\alpha > 0$, and $\gamma > 0$ are trade-off coefficients controlling the strength of the support supervision, the marginal-entropy regularization, and the text alignment, respectively.\\

The first term is the cross-entropy over the support set, anchoring the classifier to the few labeled examples. The second term estimates the mutual information between query features and predicted labels, $\hat{\mathcal{I}}_{\alpha} = \alpha\, \hat{\mathcal{H}}(Y_{\mathcal{Q}}) - \hat{\mathcal{H}}(Y_{\mathcal{Q}} | X_{\mathcal{Q}})$, simultaneously encouraging confident per-sample predictions (low conditional entropy) and balanced class marginals (high marginal entropy). The third term is a model-seeking KL divergence, $\mathcal{D}_{\mathrm{KL}}(\bm{p} \,\Vert\, \hat{\bm{y}}) = \frac{1}{|\mathcal{Q}|} \sum_{i \in \mathcal{Q}} \mathcal{D}_{\mathrm{KL}}(\bm{p}_i \,\Vert\, \hat{\bm{y}}_i)$, regularizing the optimized predictions towards the VLM's zero-shot distribution and preventing drift from pre-trained language knowledge. Problem~\eqref{eq:timpp} is solved via an Alternating Direction Method (ADM) that introduces auxiliary assignment variables $\bm{q} = [q_{ik}] \in \mathbb{R}^{|\mathcal{Q}| \times K}$ and alternates closed-form updates of $\bm{q}$ and $\mathbf{W}$ (see Sec.~\ref{sec:optim} for details).

\subsection{Main Contribution: Local Consistency Regularization}
\label{sec:lc}

\noindent \textbf{Motivation.} TIM++ operates exclusively on \emph{global} statistics of the query distribution: the marginal entropy encourages overall class balance, while the conditional entropy enforces per-sample confidence. Neither term explicitly accounts for the \emph{local} geometric structure of the query feature manifold. Yet VLM encoders produce tightly clustered representations, mapping images of the same scene class to nearby points on the unit hypersphere. A query sample's nearest neighbors are therefore very likely to share its true class. This neighborhood structure constitutes a powerful prior that comes for free at inference time, but it goes unexploited in TIM++.\\

\noindent \textbf{Formulation.} We capture this structure through a neighborhood graph over the query set. For each pair of query samples $(i,j) \in \mathcal{Q} \times \mathcal{Q}$, we define the affinity as the cosine similarity of their visual embeddings,
\begin{equation}
    a_{ij} = \bm{f}_i^{\top} \bm{f}_j ,
    \label{eq:affinity}
\end{equation}
and let $\mathcal{N}_i \subset \mathcal{Q}$ denote the $\kappa$ nearest neighbors of query $i$ under this affinity. The neighborhood-averaged prediction is
\begin{equation}
    \bar{\bm{p}}_i = \frac{1}{\kappa} \sum_{j \in \mathcal{N}_i} \bm{p}_j ,
    \label{eq:consensus}
\end{equation}
and the Local Consistency (LC) regularizer penalizes deviations of each query from its neighborhood consensus:
\begin{equation}
    \mathcal{L}_{\mathrm{LC}}(\bm{p}) = \frac{1}{|\mathcal{Q}|} \sum_{i \in \mathcal{Q}} \mathcal{D}_{\mathrm{KL}}\big(\bm{p}_i \,\Vert\, \bar{\bm{p}}_i\big) .
    \label{eq:lc}
\end{equation}
Our full objective augments TIM++ with this single additional term:
\begin{equation}
    \min_{\mathbf{W}} \;\;
    \lambda \, \mathrm{CE}(\mathbf{W}; \mathcal{S})
    \;-\; \hat{\mathcal{I}}_{\alpha}(X_{\mathcal{Q}}; Y_{\mathcal{Q}})
    \;+\; \gamma \, \mathcal{D}_{\mathrm{KL}}(\bm{p} \,\Vert\, \hat{\bm{y}})
    \;+\; \lambda_{\mathrm{LC}} \, \mathcal{L}_{\mathrm{LC}}(\bm{p}) ,
    \label{eq:ours}
\end{equation}
where $\lambda_{\mathrm{LC}} \geq 0$ controls the strength of the neighborhood regularization. Setting $\lambda_{\mathrm{LC}} = 0$ recovers TIM++ exactly.

\subsection{Multi-Source Extension: Fused Affinity Graph}
\label{sec:multisource}

\noindent \textbf{Motivation.}
The affinity of Eq.~\eqref{eq:affinity} relies solely on the VLM's CLS token, which captures global semantic similarity. However, remote sensing scene classes often share similar global appearance while differing in local texture or structural patterns. We therefore enrich the neighborhood graph with embeddings from a second encoder that captures such complementary cues, leaving the objective of Eq.~\eqref{eq:ours} untouched: the extension acts purely on how neighbors are defined.\\

\noindent \textbf{Formulation.} Let $\bm{g}_i \in \mathbb{R}^{d'}$ denote the $\ell_2$-normalized embedding of query $i$ produced by a supplemental frozen vision encoder (\emph{e.g.}, DINOv3), obtained by average-pooling the encoder's patch token embeddings. For each pair of query samples, the cosine similarities of both sources, $\tilde{s}^{\,v}_{ij} = \bm{f}_i^{\top} \bm{f}_j$ and $\tilde{s}^{\,g}_{ij} = \bm{g}_i^{\top} \bm{g}_j$, are min--max normalized to $[0,1]$ over all query pairs, yielding $s^{\,v}_{ij}$ and $s^{\,g}_{ij}$, and fused multiplicatively into a single affinity:
\begin{equation}
    a_{ij} = s^{\,v}_{ij} \cdot s^{\,g}_{ij} ,
    \label{eq:fusion}
\end{equation}
which replaces Eq.~\eqref{eq:affinity} when constructing the neighbor sets $\{\mathcal{N}_i\}_{i \in \mathcal{Q}}$. The product acts as a soft logical \emph{and}: a pair receives a high affinity only when both encoders agree, making the graph robust to faulty neighbors from a single feature space. The supplemental embeddings enter the framework only through Eq.~\eqref{eq:fusion}; all terms of the objective in Eq.~\eqref{eq:ours} remain unchanged.

\subsection{Optimization}
\label{sec:optim}

\noindent \textbf{Closed-form $\bm{q}$-update.}
We solve Eq.~\eqref{eq:ours} with the same ADM scheme as TIM++; the $\mathbf{W}$-update is unchanged. The $\bm{q}$-update is extended to incorporate the local consistency term: solving the Karush--Kuhn--Tucker (KKT) conditions of the reformulated objective under the simplex constraints $\sum_k q_{ik} = 1$, $q_{ik} \geq 0$, yields at iteration $t+1$:
\begin{equation}
    q^{(t+1)}_{ik} \;\propto\;
    \big(p^{(t)}_{ik}\big)^{1+\alpha} \cdot
    \hat{y}_{ik}^{\,\gamma} \cdot
    \big(\bar{p}^{(t)}_{ik}\big)^{\lambda_{\mathrm{LC}}} ,
    \label{eq:qupdate}
\end{equation}
followed by the renormalization $q_{ik} \leftarrow q_{ik} / \sum_{k'} q_{ik'}$.\\ 

The three factors are complementary: $(p_{ik})^{1+\alpha}$ sharpens confident predictions, $\hat{y}_{ik}^{\,\gamma}$ anchors assignments to VLM's textual knowledge, and $(\bar{p}_{ik})^{\lambda_{\mathrm{LC}}}$ reinforces predictions consistent with the neighborhood consensus. Our regularizer thus enters the solver as a single extra multiplicative factor in the closed-form update, incurring negligible computational overhead. The update is identical for the single-source and multi-source variants, which differ only in the affinity used to pre-compute the neighbor sets.\\

\noindent \textbf{kNN computation.}
The neighbor sets $\{\mathcal{N}_i\}_{i \in \mathcal{Q}}$ are computed once, before the iterative loop, from the affinity of Eq.~\eqref{eq:affinity} or its fused counterpart of Eq.~\eqref{eq:fusion}.\\

%For large query sets we compute the affinities in chunks to avoid materialising the full $|\mathcal{Q}| \times |\mathcal{Q}|$ matrix. This incurs a one-time $\mathcal{O}\big(|\mathcal{Q}|^2 (d + d')\big)$ cost; each subsequent ADM iteration then requires only $\mathcal{O}(|\mathcal{Q}| \kappa K)$ extra work to form the neighbourhood-averaged predictions of Eq.~\eqref{eq:consensus}.

%% file: sections/4-Experiments.tex
\section{Experiments}
\label{sec:exp}

\subsection{Experimental Setup}
\label{sec:expsetup}

\noindent\textbf{Datasets and splits.}
We follow the ten-dataset RS benchmark of~\cite{elkhoury2026few}:
AID~\cite{xia2017aid}, EuroSAT~\cite{helber2018eurosat}, MLRSNet~\cite{qi2020mlrsnet}, OPTIMAL31~\cite{wang2018optimal}, PatternNet~\cite{zhou2018patternnet}, RESISC45~\cite{cheng2017resisc45}, RSC11~\cite{zhao2016rsc11}, RSICB128~\cite{li2020rsicb}, RSICB256~\cite{li2020rsicb}, and WHURS19~\cite{xia2010whurs19}.
Each dataset uses a fixed 50\%/25\%/25\% train/val/test split seeded identically to~\cite{elkhoury2026few}. We evaluate $n\!\in\!\{1,2,4,8,16\}$ shots and average over 10 random
seeds.\\

\noindent\textbf{Models.}
We evaluate on two VLM backbones: standard CLIP ViT-B/32 and GeoRSCLIP ViT-B/32~\cite{zhang2024rs5m}.
Both models use the generic satellite imaging-specific text prompt template \texttt{"a centered satellite photo of \{class\}."} for their respective zero-shot prediction. For the multi-source extension, we additionally extract the mean patch token embeddings from the satellite-pretrained DINOv3 ViT-L/16~\cite{simeoni2025dinov3} (SAT-493M checkpoint), which are used exclusively at initialization to build the fused affinity graph in Eq.~\ref{eq:fusion}.\\

\noindent\textbf{Baselines.}
We compare \lctim{} against both zero-shot models and transductive few-shot methods that operate on top of frozen embeddings:
(i) zero-shot GeoRSCLIP/CLIP;
(ii) LP++~\cite{huang2024lpplusplus}, which trains a linear
probe with a learnable CLIP zero-shot weighting scalar;
(iii) TransCLIP~\cite{zanella2024boosting}, a
Gaussian-mixture transductive method with a Laplacian affinity
regularizer;
(iv) TIM++~\cite{li2026timplusplus}, the mutual-information
transductive baseline that \lctim{} extends. \\

\noindent \textbf{Hyperparameters.}
We inherit all TIM++ initialization and keep hyperparameters unchanged ($\tau = 120$, $\lambda = 0.4$, $\alpha = 1.0$, $\gamma = 0.05$, $T = 150$ iterations) and set $\kappa = 5$ and $\lambda_{\mathrm{LC}} = 0.3$.
All hyperparameters are kept fixed across all datasets and shot settings, without any dataset-specific tuning.

\subsection{Transductive Benchmarking on RS}
\label{sec:results}
We report the top-1 accuracies over the two VLMs in \Cref{tab:clip} and \Cref{tab:georsclip} at  representative shot values
$n\!\in\!\{0,1,2,4,8,16\}$ for each of the 10 datasets, enabling fine-grained
analysis for transductive methods.\\

\noindent\textbf{Transduction boosts RSVLMs on RS data.}
All four transductive methods consistently outperform zero-shot at every shot setting and on both models.
At 1-shot, the transductive methods already bring at least $+17.8\%$ on average compared to the zero-shot inductive GeoRSCLIP baseline. This transductive advantage persists at 16 shots and across both backbones, underscoring that jointly processing the query batch is more effective than independent inductive classification.\\

\noindent\textbf{\lctim{} achieves state-of-the-art.}
\lctim{} outperforms TIM++ on average at every shot level on both backbones. The gains are the biggest at 1 and 2-shot ($90.3$ vs.\ $87.7$ at 2-shot with GeoRSCLIP), where the cross-entropy signal from the support set $\mathcal{S}$ is weaker and neighborhood cues provide the most informative source of supervision. The largest per-dataset gains between TIM++ and \lctim{} for GeoRSCLIP backbone occur on RSICB128 ($+8.6\%$, 1 shot) and RSICB256 ($+6.0\%$, 2-shot), both of which have large query sets (9{,}147 and 6{,}169 images), providing dense local neighborhoods for the local-consistency term to exploit. At 16-shot, \lctim{} reaches $94.2\%$ average vs.\ $92.9\%$ for TIM++. These results confirm the benefits brought by our local-consistency regularizer in the TIM++ objective function for both low and high-shot settings.\\

% ---------------------------------------------------------------
% TABLE I — CLIP B/32
% ---------------------------------------------------------------

\input{tables/transductive_on_clip_vitb32}

% ---------------------------------------------------------------
% TABLE II — GeoRSCLIP B/32  (shots 0, 1, 4, 16)
% ---------------------------------------------------------------

\input{tables/transductive_on_georsclip_vitb32}

\newpage

% ---------------------------------------------------------------
% TABLE III — LC-TIM vs LC-TIM+D  (GeoRSCLIP, shots 1/4/16)
% ---------------------------------------------------------------

\input{tables/dino_extension}

\noindent\textbf{Transductive Method comparison.}
On GeoRSCLIP, TransCLIP peaks early (strong at 1-shot on PatternNet: $95.9\%$
and WHURS19: $97.1\%$) but plateaus after 4-shot on most datasets and can even decrease in performance on EuroSAT as the number of shots increases.
This indicates that the direct prototype learning of LP++, TIM++ and \lctim{} is more discriminative and efficient as the size of the support set grows, while TransCLIP's Gaussian formulation does not scale as much. 
We can also observe that LP++ tends to lag behind in the 1 and 2-shot settings on large datasets (AID, EuroSAT, MLRSNet, PatternNet, RESISC45) while the other methods are more efficient due to a better integration of the VLM's zero-shot prediction.

%\noindent\textbf{Effect of backbone.}
%GeoRSCLIP consistently outperforms standard CLIP, with the zero-shot gap ($+13.7\%$ on average) persisting into the few-shot regime with \lctim{} from 1-shot ($+8.4\%$) to 16-shot ($+3.5\%$). This observation is also valid for the other transductive methods, which proves that RS-specific pre-training VLMs yield representations that are both well-suited for the transductive paradigm and that are better aligned with semantic RS categories, amplifying the benefit of the local-consistency term.

\subsection{Multi-source Extension}
\label{sec:dinov3}

\noindent\textbf{Quantitative results.} \Cref{tab:lctimd} compares \lctim{} and  $\lctim{}_{\text{\tiny{+DINO}}}$ extension on GeoRSCLIP. Our extension significantly increases the average performances without adding any additional parameters to tune. While GeoRSCLIP's CLS token carries global information, DINOv3 mean patch embeddings affinities provide complementary local structure cues that improve upon \lctim{} on geometrically complex datasets with large query sets such as AID, EuroSAT, RESISC45 and MLRSNet, while remaining competitive elsewhere.
The fused affinity graph of Eq.~\ref{eq:fusion} consistently equals or improves over single-affinity \lctim{}, particularly at 1-shot where structural cues supplement the sparse semantic ones.\\

\noindent\textbf{Qualitative results.} We visualize in \Cref{fig:qualitative} the $\kappa=5$ neighbors for three sample queries in the OPTIMAL31 dataset. The example illustrates over fine-grained classes how the fused affinity recovers same-label neighbors, and thus the correct single-shot prediction, where the individual $\mathcal{N}_{\text{GeoRSCLIP}}$ graph is misled by visually similar but wrongly-labeled neighbors.

\newpage

\input{figures/qualitative1}
\newpage

% ---------------------------------------------------------------
% TABLE IV — Ablation: k and lambda
% ---------------------------------------------------------------

\input{tables/ablation}

\subsection{Ablation Study}
\label{sec:ablation}
We study the impact of the principal components involved in the \lctim{} design over four diverse datasets (AID, EuroSAT, MLRSNet and OPTIMAL31). We select the GeoRSCLIP ViT-B/32 since it outperforms CLIP in our experiments. We chose 4-shot as a balanced compromise.\\

\noindent\textbf{Impact of neighborhood size $\kappa$.}
\Cref{tab:ablation_k} (a) shows stable performances for $\kappa={1, 3, 5, 10}$ indicating low sensitivity to this parameter. However, the extreme case of a single-neighbor graph consistently degrades the performances. It means that few neighbors are enough to have steady supervision signal while keeping a sparse graph.\\

\noindent\textbf{Effect of $\lambda_\mathrm{LC}$.}
\Cref{tab:ablation_k} (b) demonstrates a low sensitivity to the value of $\lambda_\mathrm{LC}$ around its operating point ($0.3$). Again, completely removing our local-consistency term $\lambda_\mathrm{LC}=0$ is equivalent to TIM++ and translates to drops in performance. Still, there is room for dataset-specific parameter selection based on a validation set to further enhance the performances but at a greater cost.\\

\noindent\textbf{Design of the nearest-neighbors graph.} 
We investigate different choices of features for the neighborhood graph construction in \Cref{tab:ablation_k} (c) for both \lctim{} and $\lctim{}_{\text{\tiny{+DINO}}}$. With GeoRSCLIP solely, the global information carried by the CLS token is more informative than its patch tokens. This can be explained as the contrastive pre-training only learns via the CLS token and thus focuses more on global information. On the other hand, employing fine-grained DINOv3 patch tokens can either increase or decrease the performances depending on the dataset. It is when we combine both sources of information through our $\lctim{}_{\text{\tiny{+DINO}}}$ extension that we are able to capitalize on both global information and local textural cues to outperform each of the single-source approaches.\\

\noindent\textbf{Methods runtime.} \Cref{tab:ablation_k} (d) indicates that the kNN graph construction in \lctim{} only adds a negligible overhead compared to the TIM++ baseline while remaining faster than LP++ and TransCLIP.\\

%% file: tables/transductive_on_clip_vitb32.tex
% ---------------------------------------------------------------
% TABLE I --- CLIP B/32 -- lambda = 0.3
% ---------------------------------------------------------------
\begin{table}[!t]
  \centering
  \caption{Top-1 accuracy (\%) - CLIP ViT-B/32.
  \textbf{Bold} = best; \uline{underlined} = second best. Averaged over 10 random seeds.}
    \vspace{-4mm}
  \label{tab:clip}
  \setlength{\tabcolsep}{2.5pt}
  \renewcommand{\arraystretch}{1.05}
  \scriptsize
  \begin{tabularx}{\linewidth}{cl*{10}{Z}r}
    $n$ & \multicolumn{1}{l}{\textbf{Method}}
      & \rotatebox{30}{\tiny{AID}}
      & \rotatebox{30}{\tiny{EuroSAT}}
      & \rotatebox{30}{\tiny{MLRSNet}}
      & \rotatebox{30}{\tiny{OPTIMAL31}}
      & \rotatebox{30}{\tiny{PatternNet}}
      & \rotatebox{30}{\tiny{RESISC45}}
      & \rotatebox{30}{\tiny{RSC11}}
      & \rotatebox{30}{\tiny{RSICB128}}
      & \rotatebox{30}{\tiny{RSICB256}}
      & \rotatebox{30}{\tiny{WHURS19}}
      & \textbf{~~Avg} \\
    \midrule
  \textbf{0} & CLIP
    & 60.9 & 42.3 & 46.7 & 66.0 & 53.2 & 56.6 & 48.5 & 23.3 & 32.5 & 78.8 & 50.9 \\
  \midrule
  \multirow{4}{*}{\textbf{1}}
   & LP++       & 74.6 & 59.1 & 60.8 & 76.1 & 78.9 & 67.9 & \second{76.0} & 55.6 & 68.9 & 86.4 & 70.4 \\
   & TransCLIP  & 82.2 & 66.8 & 63.8 & 79.1 & 86.8 & 74.5 & \best{78.3} & 52.7 & 63.3 & \best{94.9} & 74.2 \\
   & TIM++      & \second{83.9} & \best{71.7} & \second{71.6} & \best{81.7} & \second{92.0} & \best{78.7} & 74.4 & \second{59.1} & \second{71.0} & 94.2 & \second{77.8} \\
  \rowcolor{tablelightblue}\cellcolor{white}
   & \lctim{}  & \best{86.8} & \second{71.5} & \best{74.7} & \second{79.4} & \best{93.1} & \second{78.6} & 72.8 & \best{64.4} & \best{75.5} & \second{94.5} & \best{79.1} \\
  \midrule
  \multirow{4}{*}{\textbf{2}}
   & LP++       & 81.5 & 65.1 & 65.9 & 81.2 & 83.3 & 71.6 & \best{81.2} & 66.2 & 77.4 & 91.0 & 76.4 \\
   & TransCLIP  & 85.1 & 67.4 & 64.7 & 78.3 & 89.1 & 74.8 & 77.9 & \second{67.5} & \second{79.6} & \best{95.1} & 78.0 \\
   & TIM++      & \second{86.3} & \second{72.6} & \second{74.5} & \second{81.4} & \second{93.1} & \second{79.8} & 75.5 & 66.5 & 75.1 & \second{94.7} & \second{80.0} \\
  \rowcolor{tablelightblue}\cellcolor{white}
   & \lctim{}  & \best{89.1} & \best{75.1} & \best{77.6} & \best{81.6} & \best{95.7} & \best{81.2} & \second{80.2} & \best{78.4} & \best{83.3} & \second{94.7} & \best{83.7} \\
  \midrule
  \multirow{4}{*}{\textbf{4}}
   & LP++       & 86.3 & 77.0 & 70.6 & \second{84.9} & 88.1 & 76.2 & \second{82.5} & 70.7 & 85.0 & 94.0 & 81.6 \\
   & TransCLIP  & 87.8 & 73.3 & 68.6 & 83.7 & 90.1 & 77.3 & 80.1 & \second{74.5} & \second{85.3} & \best{95.1} & 81.6 \\
   & TIM++      & \second{88.9} & \second{82.5} & \second{77.7} & \best{85.5} & \second{94.7} & \second{81.9} & 79.0 & 72.9 & 80.9 & \best{95.1} & \second{83.9} \\
  \rowcolor{tablelightblue}\cellcolor{white}
   & \lctim{}  & \best{90.5} & \best{84.4} & \best{80.3} & 83.2 & \best{96.7} & \best{82.8} & \best{84.1} & \best{83.8} & \best{91.4} & \second{94.8} & \best{87.2} \\
  \midrule
  \multirow{4}{*}{\textbf{8}}
   & LP++       & 89.0 & 80.9 & 74.5 & \best{86.9} & 90.8 & 79.3 & \best{89.0} & 74.6 & \second{87.2} & 95.0 & 84.7 \\
   & TransCLIP  & 88.5 & 73.9 & 73.2 & 84.4 & 91.8 & 79.1 & 85.8 & \second{78.9} & 84.1 & 95.7 & 83.5 \\
   & TIM++      & \second{90.1} & \second{84.4} & \second{80.1} & \second{86.5} & \second{96.0} & \second{83.2} & \second{88.1} & 78.5 & 86.3 & \best{96.5} & \second{87.0} \\
  \rowcolor{tablelightblue}\cellcolor{white}
   & \lctim{}  & \best{91.7} & \best{85.6} & \best{83.1} & 85.9 & \best{97.2} & \best{84.2} & 87.2 & \best{89.1} & \best{92.8} & \second{96.2} & \best{89.3} \\
  \midrule
  \multirow{4}{*}{\textbf{16}}
   & LP++       & 91.1 & 84.4 & 75.6 & \best{88.6} & 92.1 & 81.6 & \best{89.5} & 77.5 & 89.9 & \second{97.4} & 86.8 \\
   & TransCLIP  & 89.0 & 73.5 & 73.8 & 86.5 & 93.5 & 80.6 & 84.6 & 80.7 & 88.8 & 96.1 & 84.7 \\
   & TIM++      & \second{92.0} & \second{87.1} & \second{82.9} & \second{87.9} & \second{96.6} & \second{85.2} & 86.1 & \second{84.3} & \second{90.3} & \best{97.5} & \second{89.0} \\
  \rowcolor{tablelightblue}\cellcolor{white}
   & \lctim{}  & \best{92.6} & \best{87.2} & \best{85.0} & 87.8 & \best{97.3} & \best{85.8} & \second{88.0} & \best{92.2} & \best{94.7} & 96.4 & \best{90.7} \\
  \bottomrule
  \end{tabularx}
  \end{table}

%% file: tables/transductive_on_georsclip_vitb32.tex
% ---------------------------------------------------------------
% TABLE II — GeoRSCLIP B/32  (shots 0, 1, 4, 16) -- lambda = 0.3
% ---------------------------------------------------------------
\begin{table}[!t]
  \centering

  \caption{Top-1 accuracy (\%) - GeoRSCLIP ViT-B/32.
  \textbf{Bold} = best; \uline{underlined} = second best. Averaged over 10 random seeds.}
    \vspace{-4mm}
  \label{tab:georsclip}
  \setlength{\tabcolsep}{2.5pt}
  \renewcommand{\arraystretch}{1.05}
  \scriptsize
  \begin{tabularx}{\linewidth}{cl*{10}{Z}r}
    $n$ & \multicolumn{1}{l}{\textbf{Method}}
      & \multicolumn{1}{c}{\rotatebox{30}{\tiny{AID}}}
      & \rotatebox{30}{\tiny{EuroSAT}}
      & \rotatebox{30}{\tiny{MLRSNet}}
      & \rotatebox{30}{\tiny{OPTIMAL31}}
      & \rotatebox{30}{\tiny{PatternNet}}
      & \rotatebox{30}{\tiny{RESISC45}}
      & \rotatebox{30}{\tiny{RSC11}}
      & \rotatebox{30}{\tiny{RSICB128}}
      & \rotatebox{30}{\tiny{RSICB256}}
      & \rotatebox{30}{\tiny{WHURS19}}
      & \textbf{~~Avg} \\
    \midrule

  \textbf{0} & GeoRSCLIP
    & 70.6 & 52.7 & 64.2 & 78.9 & 77.8 & 70.9 & 65.6 & 28.9 & 48.2 & 87.9 & 64.6 \\
  \midrule
  \multirow{4}{*}{\textbf{1}}
   & LP++       & 83.4 & 72.5 & 72.9 & 86.4 & 90.7 & 79.2 & \best{89.0} & \second{73.3} & \second{81.3} & 94.9 & 82.4 \\
   & TransCLIP  & \best{91.6} & 80.3 & 77.6 & \second{87.6} & 95.9 & 85.1 & \second{86.7} & 69.6 & 79.1 & \best{97.1} & 85.1 \\
   & TIM++      & 87.8 & \best{87.0} & \second{79.2} & 87.4 & \second{96.7} & \second{86.1} & \second{86.7} & 67.8 & 80.0 & \best{97.1} & \second{85.6} \\
  \rowcolor{tablelightblue}\cellcolor{white}
   & \lctim{}  & \second{89.6} & \second{86.1} & \best{82.5} & \best{88.6} & \best{97.6} & \best{87.8} & 85.9 & \best{76.4} & \best{83.4} & \second{96.9} & \best{87.5} \\
  \midrule
  \multirow{4}{*}{\textbf{2}}
   & LP++       & 90.2 & 78.3 & 76.9 & 89.5 & 92.9 & 83.0 & 87.3 & \second{82.9} & \best{89.0} & 96.5 & 86.7 \\
   & TransCLIP  & \second{92.2} & 81.7 & 77.3 & 88.8 & 96.0 & 86.0 & \second{87.9} & 82.7 & 88.3 & 96.5 & \second{87.7} \\
   & TIM++      & 90.4 & \second{87.7} & \second{81.3} & \second{89.8} & \second{96.6} & \second{87.0} & 87.3 & 77.5 & 82.6 & \second{96.9} & \second{87.7} \\
  \rowcolor{tablelightblue}\cellcolor{white}
   & \lctim{}  & \best{92.5} & \best{89.4} & \best{85.4} & \best{90.8} & \best{97.6} & \best{88.4} & \best{88.1} & \best{85.4} & \second{88.6} & \best{97.1} & \best{90.3} \\
  \midrule
  \multirow{4}{*}{\textbf{4}}
   & LP++       & 92.1 & 84.0 & 81.0 & \second{91.5} & 95.1 & 86.5 & 88.0 & \second{88.1} & \best{92.3} & \best{97.5} & 89.6 \\
   & TransCLIP  & \second{92.6} & 82.3 & 79.7 & 91.3 & 95.4 & 86.5 & 88.1 & 87.0 & 89.7 & 96.5 & 88.9 \\
   & TIM++      & 91.4 & \second{89.7} & \second{84.1} & 91.3 & \second{96.7} & \second{87.9} & \best{90.3} & 81.7 & 87.9 & \second{97.4} & \second{89.8} \\
  \rowcolor{tablelightblue}\cellcolor{white}
   & \lctim{}  & \best{94.1} & \best{92.0} & \best{86.7} & \best{91.6} & \best{97.7} & \best{89.3} & \second{89.9} & \best{88.8} & \second{92.0} & 97.0 & \best{91.9} \\
  \midrule
  \multirow{4}{*}{\textbf{8}}
   & LP++       & \second{93.5} & 89.7 & 83.6 & 92.1 & 96.2 & 87.5 & \best{93.9} & \second{89.5} & \second{93.4} & \best{98.3} & \second{91.8} \\
   & TransCLIP  & 92.9 & 81.0 & 82.6 & 92.0 & 96.4 & 87.4 & 89.0 & 87.8 & 90.6 & 97.5 & 89.7 \\
   & TIM++      & 92.9 & \second{91.4} & \second{85.9} & \second{92.4} & \second{97.5} & \second{88.8} & \best{93.9} & 85.4 & 90.0 & \second{97.6} & 91.6 \\
  \rowcolor{tablelightblue}\cellcolor{white}
   & \lctim{}  & \best{94.6} & \best{93.1} & \best{88.5} & \best{92.8} & \best{98.1} & \best{90.2} & \second{91.4} & \best{92.3} & \best{94.0} & 97.4 & \best{93.2} \\
  \midrule
  \multirow{4}{*}{\textbf{16}}
   & LP++       & \second{94.9} & 90.0 & 85.0 & \best{93.2} & 96.9 & 88.0 & \best{93.8} & \second{90.5} & \second{95.4} & \best{99.0} & 92.7 \\
   & TransCLIP  & 93.6 & 80.9 & 82.3 & 92.7 & 96.3 & 87.8 & 88.3 & 89.0 & 93.4 & 96.7 & 90.1 \\
   & TIM++      & 94.0 & \second{92.0} & \second{87.7} & \second{93.0} & \second{97.7} & \second{90.6} & \second{93.7} & 90.2 & 93.1 & \second{97.5} & \second{92.9} \\
  \rowcolor{tablelightblue}\cellcolor{white}
   & \lctim{}  & \best{95.8} & \best{93.4} & \best{90.0} & \best{93.2} & \best{98.3} & \best{91.1} & 92.3 & \best{95.2} & \best{95.6} & 97.3 & \best{94.2} \\
  \bottomrule
  \end{tabularx}
  \end{table}

%% file: tables/dino_extension.tex
\begin{table}[!t]
\centering
\caption{\lctim{} vs.\ $\lctim{}_{\text{\tiny{+DINO}}}$ (GeoRSCLIP ViT-B/32 + DINOv3
ViT-L/16 SAT-493M). \textbf{Bold} = best. Top-1 accuracy averaged over 10 random seeds.}
  \vspace{-3mm}
\label{tab:lctimd}
\setlength{\tabcolsep}{2.5pt}
\renewcommand{\arraystretch}{1.05}
\scriptsize
\begin{tabularx}{\linewidth}{cl*{10}{Z}r}
    $n$ & \multicolumn{1}{l}{\textbf{Method}}
      & \multicolumn{1}{c}{\rotatebox{30}{\tiny{AID}}}
      & \rotatebox{30}{\tiny{EuroSAT}}
      & \rotatebox{30}{\tiny{MLRSNet}}
      & \rotatebox{30}{\tiny{OPTIMAL31}}
      & \rotatebox{30}{\tiny{PatternNet}}
      & \rotatebox{30}{\tiny{RESISC45}}
      & \rotatebox{30}{\tiny{RSC11}}
      & \rotatebox{30}{\tiny{RSICB128}}
      & \rotatebox{30}{\tiny{RSICB256}}
      & \rotatebox{30}{\tiny{WHURS19}}
      & \textbf{~~Avg} \\
    \midrule
\multirow{2}{*}{\textbf{1}}
& \lctim{}  & 89.6 & 86.1 & 82.5 & 88.6 & 97.6 & 87.8 & 85.9 & 76.4 & 83.4 & 96.9 & 87.5 \\
%\rowcolor{tablelightblue}\cellcolor{white}
&\cellcolor{tablelightblue}$\lctim{}_{\text{\tiny{+DINO}}}$ &\cellcolor{tablelightblue}\best{89.9} &\cellcolor{tablelightblue}\best{90.2} &\cellcolor{tablelightblue}\best{83.2} &\cellcolor{tablelightblue}\best{90.0} &\cellcolor{tablelightblue}\best{98.0} & \cellcolor{tablelightblue}\best{88.3} &\cellcolor{tablelightblue}\best{86.7} &\cellcolor{tablelightblue}\best{76.7} &\cellcolor{tablelightblue}\best{83.8} &\cellcolor{tablelightblue}\best{97.1} &\cellcolor{tablelightblue}\best{88.4} \\
\midrule
\multirow{2}{*}{\textbf{2}}
& \lctim{}  & 92.5 & 89.4 & 85.4 & 90.8 & 97.6 & 88.4 & 88.1 & 85.4 & 88.6 & 97.1 & 90.3 \\
%\rowcolor{tablelightblue}\cellcolor{white}
&\cellcolor{tablelightblue}$\lctim{}_{\text{\tiny{+DINO}}}$ & \cellcolor{tablelightblue}\best{93.1} &\cellcolor{tablelightblue}\best{91.2} &\cellcolor{tablelightblue}\best{86.2} &\cellcolor{tablelightblue}\best{91.8} &\cellcolor{tablelightblue}\best{98.0} &\cellcolor{tablelightblue}\best{89.1} &\cellcolor{tablelightblue}\best{88.3} &\cellcolor{tablelightblue}\best{85.7} &\cellcolor{tablelightblue}\best{88.8} &\cellcolor{tablelightblue}\best{97.2} &\cellcolor{tablelightblue}\best{91.0} \\
\midrule
\multirow{2}{*}{\textbf{4}}
& \lctim{}  & 94.1 & 92.0 & 86.7 & 91.6 & 97.7 & 89.3 & \best{89.9} & 88.8 & 92.0 & 97.0 & 91.9 \\
%\rowcolor{tablelightblue}\cellcolor{white}
&\cellcolor{tablelightblue}$\lctim{}_{\text{\tiny{+DINO}}}$ & \cellcolor{tablelightblue} \best{94.8} & \cellcolor{tablelightblue} \best{93.6} & \cellcolor{tablelightblue} \best{87.7} & \cellcolor{tablelightblue} \best{93.1} & \cellcolor{tablelightblue} \best{98.2} & \cellcolor{tablelightblue} \best{90.1} & \cellcolor{tablelightblue} 89.6 & \cellcolor{tablelightblue} \best{89.4} & \cellcolor{tablelightblue} \best{92.3} & \cellcolor{tablelightblue} \best{97.3} & \cellcolor{tablelightblue} \best{92.6} \\
\midrule
\multirow{2}{*}{\textbf{8}}
& \lctim{}  & 94.6 & 93.1 & 88.5 & 92.8 & 98.1 & 90.2 & \best{91.4} & 92.3 & \best{94.0} & \best{97.4} & 93.2 \\
%\rowcolor{tablelightblue}\cellcolor{white}
&\cellcolor{tablelightblue}$\lctim{}_{\text{\tiny{+DINO}}}$ & \cellcolor{tablelightblue} \best{95.0} & \cellcolor{tablelightblue} \best{94.0} & \cellcolor{tablelightblue} \best{89.4} & \cellcolor{tablelightblue} \best{94.5} & \cellcolor{tablelightblue} \best{98.5} & \cellcolor{tablelightblue} \best{90.8} & \cellcolor{tablelightblue} 91.0 & \cellcolor{tablelightblue} \best{92.9} & \cellcolor{tablelightblue} 93.9 & \cellcolor{tablelightblue} \best{97.4} & \cellcolor{tablelightblue} \best{93.7} \\
\midrule
\multirow{2}{*}{\textbf{16}}
& \lctim{}  & 95.8 & 93.4 & 90.0 & 93.2 & 98.3 & 91.1 & \best{92.3} & 95.2 & 95.6 & 97.3 & 94.2 \\
%\rowcolor{tablelightblue}\cellcolor{white}
&\cellcolor{tablelightblue}$\lctim{}_{\text{\tiny{+DINO}}}$ & \cellcolor{tablelightblue} \best{96.2} & \cellcolor{tablelightblue} \best{94.1} & \cellcolor{tablelightblue} \best{90.8} & \cellcolor{tablelightblue} \best{95.0} & \cellcolor{tablelightblue} \best{98.8} & \cellcolor{tablelightblue} \best{91.7} & \cellcolor{tablelightblue} 92.2 & \cellcolor{tablelightblue} \best{95.7} & \cellcolor{tablelightblue} \best{95.9} & \cellcolor{tablelightblue} \best{97.4} & \cellcolor{tablelightblue} \best{94.8} \\
\bottomrule
\end{tabularx}
\end{table}

%% file: figures/qualitative1.tex
\begin{figure}[!t]

\centering
\includegraphics[width=0.92\textwidth,trim=0cm 0cm 1.9cm 0cm,clip]{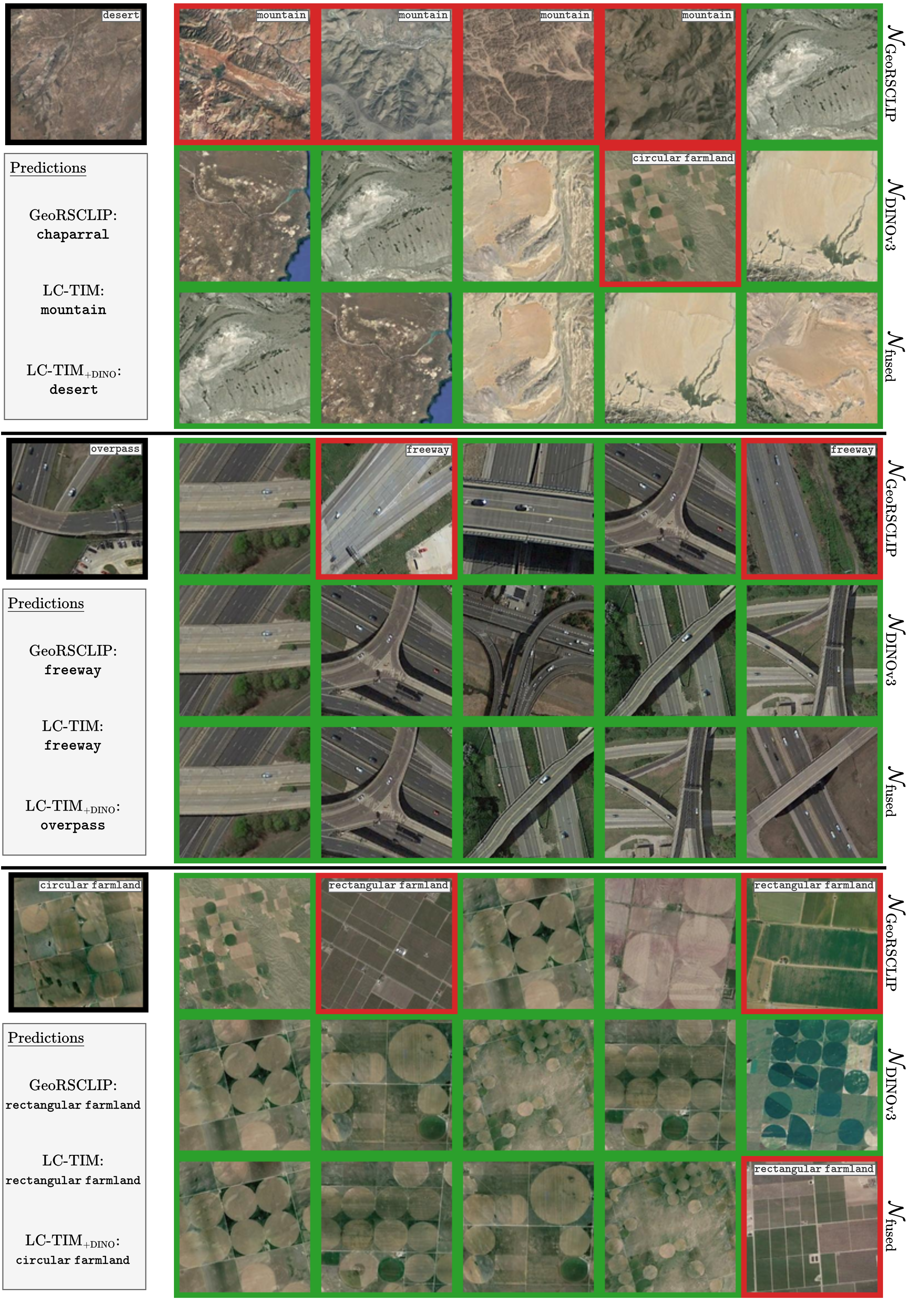}
\caption{Qualitative neighborhoods comparison between GeoRSCLIP, DINOv3 mean patch embedding and our proposed fused approach on 3 queries from OPTIMAL31. For each query, the 3 rows show the $\kappa=5$ closest neighbors according to each kNN graph $\mathcal{N}$. \lctim{} and $\lctim{}_{\text{\tiny{+DINO}}}$ use a single shot for their prediction. Green outline indicates that the neighbor has the same ground truth label as the query image.}
\label{fig:qualitative}
\end{figure}

%% file: tables/ablation.tex
\begin{table}[!h]
\centering
\caption{Ablation study of \lctim~over GeoRSCLIP ViT-B/32 at 4-shot.}
\vspace{-5mm}
\label{tab:ablation_k}
\setlength{\tabcolsep}{4pt}
\small
\begin{subtable}[t]{0.47\linewidth}
\centering
\caption{Neighborhood size $\kappa$.}
\begin{tabular}{ccccc}
\toprule
$\kappa$ & \multicolumn{1}{c}{\rotatebox{0}{\tiny{AID}}} & \rotatebox{0}{\tiny{EuroSAT}} & \rotatebox{0}{\tiny{MLRSNet}} & \rotatebox{0}{\tiny{OPT31}} \\
\midrule
1 & 92.0 & 89.7 & 86.7 & 89.7 \\
3 & 94.0 & 91.4 & 87.0 & 91.4 \\
\rowcolor{tablelightblue}
5 & 94.2 & 91.5 & 87.0 & 91.2 \\
10 & 94.1 & 91.3 & 86.9 & 92.0 \\
%20 & \best{94.4} & 91.0 & 86.7 & 90.9 \\
\bottomrule
\end{tabular}
\end{subtable}
\hfill
\begin{subtable}[t]{0.47\linewidth}
\centering
\caption{Local-consistency parameter $\lambda_\mathrm{LC}$.}
\begin{tabular}{ccccc}
\toprule
$\lambda_\mathrm{LC}$ & \multicolumn{1}{c}{\rotatebox{0}{\tiny{AID}}} & \rotatebox{0}{\tiny{EuroSAT}} & \rotatebox{0}{\tiny{MLRSNet}} & \rotatebox{0}{\tiny{OPT31}} \\
\midrule
0.0 & 91.4 & 89.7 & 84.1 & 91.3 \\
0.1 & 93.7 & 91.0 & 86.1 & 92.3 \\
%0.2 & 94.1 & 91.8 & 86.8 & 91.8 \\
\rowcolor{tablelightblue}
0.3 & 94.2 & 91.5 & 87.0 & 91.2 \\
%0.4 & 94.2 & 90.7 & 86.9 & 90.9 \\
0.5 & 94.0 & 90.3 & 86.9 & 90.8 \\
\bottomrule
\end{tabular}
\end{subtable}
\\
\vspace{6 mm}
\begin{subtable}[t]{0.47\linewidth}
\centering

\caption{Feature source for the kNN graph. \texttt{[CLS]}: class token.\texttt{[p]}: mean patch token.}
\begin{tabular}{cc|cccc}
\toprule
\begin{comment}
Features & AID & EurS & MLRS & OPT \\
\midrule
\rowcolor{tablelightblue}
GeoRS\texttt{[CLS]} & 94.2 & 91.5 & 87.0 & 91.2 \\
GeoRS\texttt{[patch]} & 93.6 & 91.2 & 86.5 & 91.5 \\
DINO\texttt{[CLS]} & 88.1 & 92.5 & 86.5 & 84.2 \\
DINO\texttt{[patch]} & 91.9 & 93.4 & 87.6 & 89.9 \\
\rowcolor{tablelightblue}  
\makecell[l]{GeoRS\texttt{[CLS]}\\ +DINO\texttt{[patch]}} & 95.1 & 93.4 & 88.1 & 92.3 \\
\end{comment}

GeoRS & DINO & \multicolumn{1}{c}{\rotatebox{0}{\tiny{AID}}} & \rotatebox{0}{\tiny{EuroSAT}} & \rotatebox{0}{\tiny{MLRSNet}} & \rotatebox{0}{\tiny{OPT31}} \\
\midrule
\rowcolor{tablelightblue}
\texttt{[CLS]} & \xmark & 94.2 & 91.5 & 87.0 & 91.2 \\
\texttt{[p]} & \xmark & 93.6 & 91.2 & 86.5 & 91.5 \\
\xmark & \texttt{[CLS]} & 88.1 & 92.5 & 86.5 & 84.2 \\
\xmark & \texttt{[p]} & 91.9 & 93.4 & 87.6 & 89.9 \\
\rowcolor{tablelightblue}  
\texttt{[CLS]} & \texttt{[p]} & 95.1 & 93.4 & 88.1 & 92.3 \\

\bottomrule
\end{tabular}
\end{subtable}
\hfill
\begin{subtable}[t]{0.47\linewidth}
\centering

\caption{Runtime on EuroSAT (8{,}100
queries, single GPU). Feature extraction excluded.}
\begin{tabular}{lcc}
\toprule
Method & Acc. & Time (s) \\
\midrule
Zero-shot & 52.7 & <0.01 \\
LP++ & 84.0 & 0.34 \\
TransCLIP & 82.3 & 0.51 \\
TIM++ & 89.7 & 0.22 \\
\rowcolor{tablelightblue}
LC-TIM & 91.5 & 0.32 \\ 
\bottomrule
\end{tabular}
\end{subtable}

\end{table}

%% file: sections/5-Conclusion.tex
\section{Conclusion}
\label{sec:conclusion}

We presented $\lctim$, a transductive few-shot adaptation method for remote sensing scene classification that augments the TIM++ objective with a local consistency regularizer. By encouraging each query prediction to agree with its nearest feature-space neighbors, our approach exploits the local geometric structure of the query manifold, a prior that comes for free at inference time. We additionally introduced $\lctim{}_{\text{\tiny{+DINO}}}$, which enriches the neighborhood graph by fusing RSVLM and DINOv3 satellite patch embeddings, providing complementary local textural cues that yield further gains.\\

Through the first comprehensive benchmark for transductive few-shot RS scene classification, spanning ten datasets, two backbones, and five shot settings, we demonstrated that transduction substantially outperforms zero-shot inductive inference on RS data, and that LC-TIM consistently surpasses competing transductive methods even under domain shifts. The improvements are most pronounced in the low-shot regime, where support supervision is weakest and neighborhood consistency is most valuable. Note that our core formulation is domain-agnostic and directly applicable to natural image few-shot benchmarks; its particular relevance to RS stems from the multi-source extension, which leverages DINOv3's satellite-pretrained variant, the best state-of-the-art self-supervised RS image encoder to date. We hope that our open-source benchmark and method encourage further study of transductive adaptation as a practical paradigm for the batch-oriented inference pipelines characteristic of remote sensing. Future work includes extending the local consistency formulation to additional structural priors, exploring adaptive neighborhood sizes, and validating the approach on operational, large-scale satellite imagery as well as classical natural-image benchmarks.